\documentclass{article}

\PassOptionsToPackage{table}{xcolor}

\usepackage[final,dblblindworkshop]{neurips_2026}
\workshoptitle{UserSim}

\makeatletter
\renewcommand{\@noticestring}{\@workshoptitle\ Workshop at NeurIPS \@neuripsyear.}
\makeatother

\usepackage[utf8]{inputenc} %
\usepackage[T1]{fontenc}    %
\usepackage{hyperref}       %
\usepackage{url}            %
\usepackage{booktabs}       %
\usepackage{amsfonts}       %
\usepackage{nicefrac}       %
\usepackage{microtype}      %
\usepackage{xcolor}         %

\usepackage{latexsym}
\usepackage{amsmath}
\usepackage{amssymb}
\usepackage{tabularx}
\usepackage{float}
\usepackage{multirow}
\usepackage{xspace}
\usepackage{inconsolata}
\usepackage{graphicx}
\usepackage{wrapfig}
\usepackage[most]{tcolorbox}
\usepackage{enumitem}

\graphicspath{{../acl/}}

\newcommand{\llamaseventy}{Llama 3.3 70B\xspace}
\newcommand{\qwenthirtytwo}{Qwen3 32B\xspace}

\tcbset{
  enhanced jigsaw,
  breakable,
  top=6pt,
  bottom=6pt,
  left=6pt,
  right=6pt,
  colback=black!5,
  colframe=black!80,
  boxrule=0.8pt,
  arc=2pt,
}

\newtcolorbox{goalboxwide}{
  colback=black!5,
  colframe=black!80,
  boxrule=0.8pt,
  arc=2pt,
  breakable,
  enhanced jigsaw,
  title={User Simulator Goal},
  fonttitle=\bfseries,
  center title
}

\ifdefined\showcomments
    \newcommand{\jt}[1]{\textcolor{red}{$_{Jayasimha}${[#1]}}}
    \newcommand{\mf}[1]{\textcolor{blue}{$_{MF}${[#1]}}}
\else
    \newcommand{\jt}[1]{}
    \newcommand{\mf}[1]{}
\fi

\title{Reliable User Simulation in Complex Multi-Step Workflows}

\renewcommand*{\thefootnote}{\fnsymbol{footnote}}

\author{%
  Pusala Pavan Teja\textsuperscript{1}\thanks{Work done by first author during an internship at Amazon.} \quad
  Jayasimha Talur\textsuperscript{2}\thanks{Corresponding author: \texttt{talurj@amazon.com}} \quad
  Saab Mansour\textsuperscript{2} \quad
  Marcello Federico\textsuperscript{2} \\
  \textsuperscript{1}IIT Palakkad \qquad \textsuperscript{2}Amazon \\
}

\begin{document}

\maketitle

\renewcommand*{\thefootnote}{\arabic{footnote}}
\setcounter{footnote}{0}

\begin{abstract}
LLM-based user simulators are increasingly used to evaluate conversational assistants. Yet, these simulators can drift from their assigned goals or fail to exhaustively explore all specified options, causing evaluation errors incorrectly attributed to the assistant. Prior work has examined goal alignment in single-task settings, but many practical scenarios require simulators to execute multi-step workflows with sequential tasks, fallback logic, and cross-task dependencies, where simulator fidelity has not been measured. To close this gap, we introduce a benchmark of 500 workflows represented using the Hierarchical Task Network (HTN) abstraction.
Our experiments reveal that the simulator's fidelity degrades sharply with workflow complexity, and that goal drift accounts for the majority of failures. Fidelity also depends on the assistant's response style, with the same simulator scoring substantially better or worse across styles.
To address this, we propose ReUST (Reasoning~+~User State Tracking), a simple prompting strategy to maintain state information across dialogue turns. ReUST improves goal success rate by more than 12 points for mid-sized and large models, and reduces simulator sensitivity to the assistant's response style.
\end{abstract}  

\section{Introduction}

\begin{wrapfigure}{r}{0.5\textwidth}
  \vspace{-\intextsep}
  \centering
  \includegraphics[width=\linewidth]{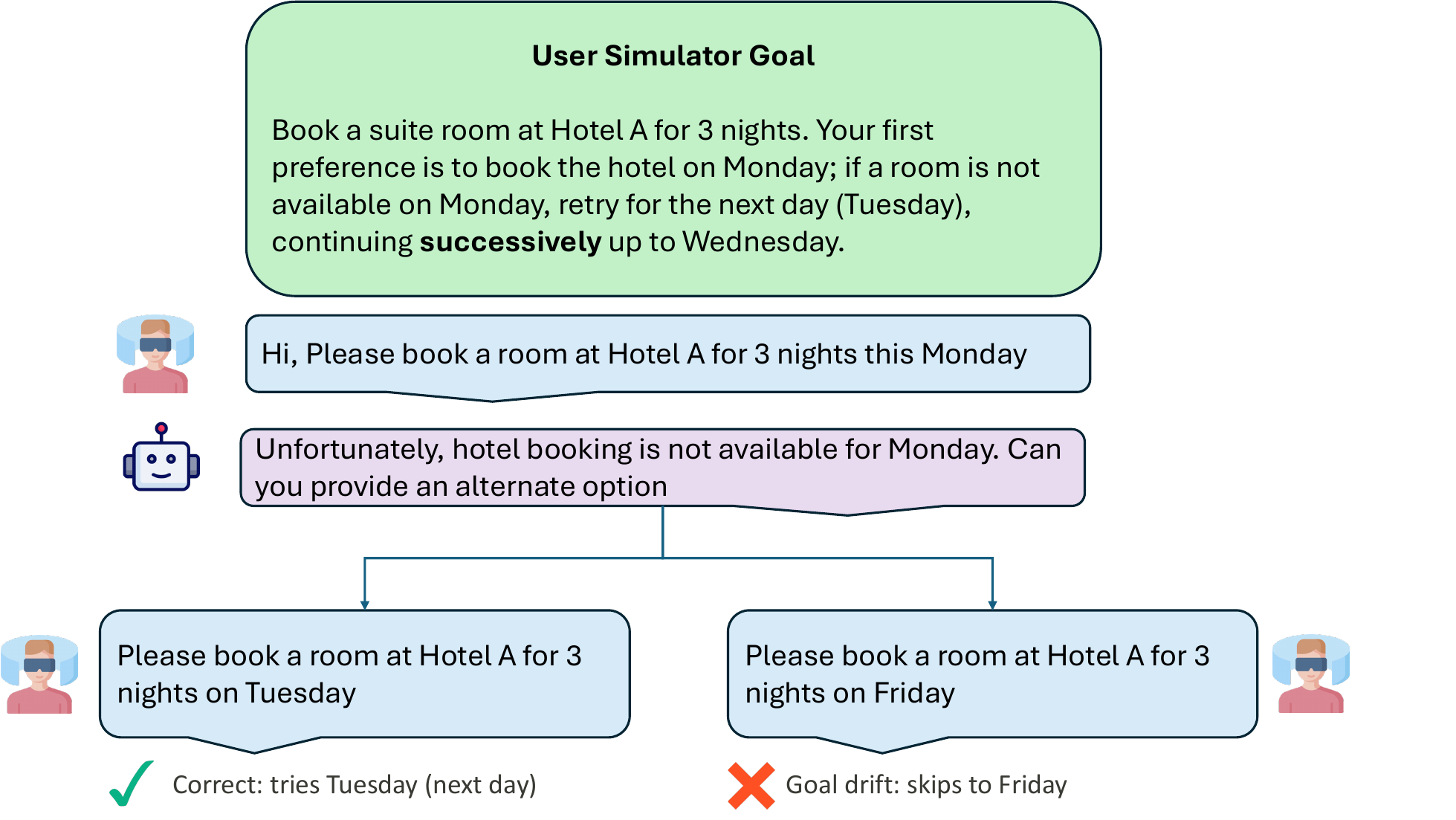}
  \caption{Goal drift: the goal instructs successive day retries (Monday $\rightarrow$ Tuesday $\rightarrow$ Wednesday), but after Monday fails the simulator skips to Friday, violating the specified ordering.}
  \label{fig:user-simulation-error}
\end{wrapfigure}
LLM-based user simulators are increasingly used to evaluate tool-augmented conversational assistants at scale \citep{yao2024taubenchbenchmarktoolagentuserinteraction,barres2025tau2benchevaluatingconversationalagents}. An LLM plays the role of a user, talking to an assistant to test whether it can book a hotel, schedule an appointment, or handle a customer service request. These frameworks rest on the assumption that the simulator faithfully follows its assigned goal. When it does not, and instead drifts from the goal, omits required information, or invents constraints, its mistakes appear as assistant errors. The failure also reduces coverage; a simulator that abandons a goal before trying the remaining options never steers the conversation to the edge cases the evaluation was designed to check. Figure~\ref{fig:user-simulation-error} shows the simulator drifting from the retry sequence its goal specified.

Recent work has begun to examine user simulator reliability (\S\ref{sec:related}), proposing methods such as goal decomposition \citep{mehri2025goalalignmentllmbaseduser}, dual-LLM verification \citep{luo2024duetsimbuildingusersimulator}, and fine-tuning on real dialogue data \citep{sekulić2024reliablellmbasedusersimulator}. These approaches evaluate simulators only on single-domain dialogues with a single overall goal. Many practical applications require simulators to execute multi-step workflows, which are structured sequences of tasks with fallback logic, retry behavior, and cross-task dependencies. For example, to test a travel booking workflow, user simulator might require completing hotel reservation, then train booking (with taxi as fallback if trains are unavailable), then restaurant reservation---where the restaurant booking time depends on the train arrival time. No existing benchmark systematically evaluates whether simulators can maintain goal alignment across such complex workflows.

We address this gap with two contributions. First, we introduce a 500-workflow evaluation set represented using the Hierarchical Task Network (HTN) abstraction \citep{Nau_2003}. Workflows vary along two axes of difficulty: \emph{path length}, the number of steps the simulator must attempt before the workflow completes, and \emph{fallback depth}, the number of tasks whose preferred method fails outright and forces a switch to an alternative. Second, we propose ReUST (Reasoning + User State Tracking), a task-agnostic prompting strategy that requires the simulator to maintain a structured reasoning state across dialogue turns, tracking which options it has tried, what the previous turn returned, and what to do next. We isolate the contribution of each component by ablation (\S\ref{sec:ablation-study}).

We measure Goal Success Rate (GSR), which is the fraction of workflows a simulator completes as prescribed in the goal. Our first finding is that GSR is not a property of the simulator alone: holding the simulator fixed and changing only how the assistant words its replies changes GSR from 62.8\% to 92.8\%, making the assistant's response style a substantial confounder for measuring fidelity (\S\ref{sec:persona-sweep}). Our proposed ReUST prompt improves GSR across assistant style making it a practical tool for faithful workflow following user simulators.

\section{Related Work}
\label{sec:related}

\paragraph{User Simulators.} LLM-based user simulators are now central to scalable dialogue evaluation \citep{davidson2023usersimulationlargelanguage,ni2026surveyllmconversationalusersimulation}. DuetSim \citep{luo2024duetsimbuildingusersimulator} adds a verifier LLM to catch simulator errors and hallucinations, DAUS \citep{sekulić2024reliablellmbasedusersimulator} fine-tunes on real dialogue data to improve goal adherence, and UGST \citep{mehri2025goalalignmentllmbaseduser}, closest to our work, decomposes user goals into profile, policy, and task components tracked across turns to prevent goal drift, while ChatChecker \citep{mayr2025chatcheckernoncooperativeusersimulation} probes robustness through non-cooperative user simulation. These methods evaluate single-task scenarios. We extend the question to multi-task workflows that require sequential execution, fallback logic, and cross-task argument propagation.

\paragraph{Dialogue State Tracking and Reasoning.} ReUST's per-turn state block is closely related to Dialogue State Tracking (DST), which maintains user goals and slot values across turns in task-oriented dialogue. Recent work recasts DST as an LLM task: driving it directly with LLMs \citep{feng2023llmdrivendialoguestatetracking}, framing it as zero-shot function calling \citep{li2024llmzeroshotdialoguestate}, and showing zero-shot LLMs performing state tracking and API calls, though still short of specialized models \citep{hudecek2023llmalltaskorienteddialogue}. ReUST differs in role and direction: rather than tracking the \emph{assistant's} belief state to satisfy a user goal, the simulator tracks its \emph{own} progress through an assigned workflow. Its structured Thought block follows the reasoning-and-acting lineage of ReAct \citep{yao2023reactsynergizingreasoningacting}, externalizing working memory so the model need not reconstruct it from the transcript at each turn.

\paragraph{Benchmarks and Datasets.} Earlier task-oriented dialogue corpora are static: MultiWOZ \citep{budzianowski2020multiwozlargescalemultidomain} covers hotel, train, taxi, restaurant, and attraction bookings, and the Schema-Guided Dialogue dataset \citep{rastogi2020scalablemultidomainconversationalagents} extends this to 20 domains with explicit schemas, but both consist of pre-recorded conversations. Interactive benchmarks followed. $\tau$-Bench \citep{yao2024taubenchbenchmarktoolagentuserinteraction} measures task success and consistency across repeated trials under domain policies, and $\tau^2$-Bench \citep{barres2025tau2benchevaluatingconversationalagents} extends it to a dual-control setting in which the simulated user also executes tools on the shared environment state. ToolSandbox \citep{lu2025toolsandboxstatefulconversational} adds stateful tool execution and a built-in user simulator for on-policy conversational evaluation. CONFETTI \citep{alkhouli2025confetticonversationalfunctioncallingevaluation} evaluates function calling turn by turn. Static or interactive, all of these measure \emph{assistant} capability while assuming faithful user simulation. We turn the measurement on the simulator, asking whether it executes its assigned workflow at all, on dynamic workflows with systematic variation in path length, fallback depth, and cross-task dependencies. Our finding that fidelity degrades with workflow complexity matches work on long-horizon tasks \citep{kwa2025measuringaiabilitycomplete}, where reliability falls as task duration grows.

\section{Benchmark Construction}

\begin{figure*}[!ht]
  \centering
  \includegraphics[width=\textwidth]{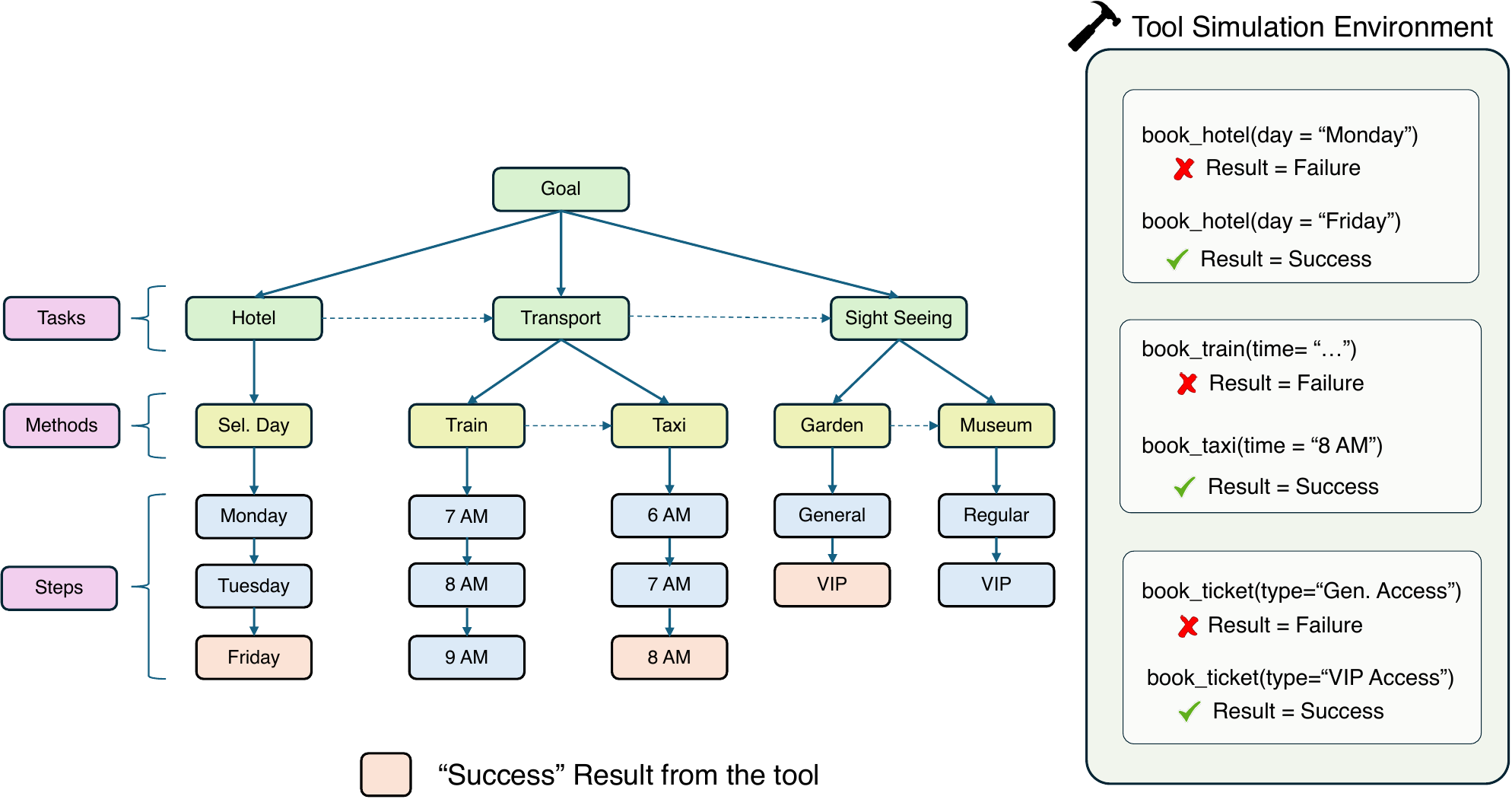}
  \caption{Hierarchical workflow structure and tool simulation environment. \textbf{Left:} the workflow forms an OR-tree over three levels (tasks, methods, steps); tasks run sequentially while methods and steps are attempted depth-first until one succeeds. Highlighted nodes are the selected execution path, 11 steps with one fallback. \textbf{Right:} the API calls along that path, abbreviated to the first failure and the final success of each task.}
  \label{fig:workflow-structure}
\end{figure*}

We represent workflows using the Hierarchical Task Network (HTN) abstraction \citep{Nau_2003,Langley2025LearningHT,Siddiqui2024HTNBasedTA}. HTNs allow arbitrary nesting, but we fix a three-level hierarchy of tasks, methods, and steps, which is enough to capture the sequential execution, fallback logic, and retry behavior of real workflows. Each workflow is serialized into deterministic natural-language instructions, and the simulator must decide at each turn whether to retry with the next step, fall back to another method, or advance to the next task.

\subsection{Workflow Structure}

A workflow is a sequence of \textbf{tasks} that must be completed in order (Figure~\ref{fig:workflow-structure}). A task is a major unit of work such as Hotel Booking or Inter-city Travel, and the simulator cannot move to the next task until the current one succeeds. Each task can be completed by one or more \textbf{methods}, attempted in sequence: Inter-city Travel might first try Train and fall back to Taxi once every train option fails. Methods correspond to the tools available to the assistant. Each method is an ordered list of \textbf{steps} tried one at a time until one succeeds, where a step is a single action with concrete parameter values, such as booking a 7 AM train and then 8 AM if that fails. The entities in a step correspond to tool arguments.

\subsection{Execution Path}

The simulator receives the workflow as a goal and must navigate it through conversation. At each step it makes a request, the assistant executes a tool call, and the simulator learns the outcome only from the assistant's response. It must then decide whether to advance to the next task, retry with the next step, or switch to a fallback method because the current one is exhausted. The \textbf{execution path} is the sequence of steps attempted from start to successful completion. Figure~\ref{fig:workflow-structure} shows a three-task workflow and one such path: the hotel task succeeds on its third step, the train method is exhausted so transport falls back to taxi, which succeeds on its own third step, and sightseeing succeeds on the second step of its primary method, giving a path of length 11 with one fallback.

\subsection{Cross-Task Dependencies}

Workflows also contain \textbf{cross-task dependencies}, where the outcome of one task fixes the parameters of a later one. These take three forms: \emph{direct propagation}, reusing a value from a completed task (``book a taxi on the same day the hotel was reserved''); \emph{computed derivation}, calculating a value from a prior outcome (``book a taxi 1 hour after the train arrival time''); and \emph{conditional branching}, where which value applies depends on how an earlier task resolved (``if the train was booked, book sightseeing 1 hour after train arrival; if the taxi was used instead, 2 hours after taxi arrival''). Handling them requires the simulator to keep a coherent internal state and transform information across task boundaries. Appendix~\ref{sec:example-goal} shows a full simulator goal.

\subsection{Dataset}
\label{sec:dataset}

\begin{wraptable}{r}{0.40\textwidth}
\vspace{-\intextsep}
\centering
\caption{Dataset overview for the 500-workflow stratified evaluation set used in all experiments.}
\label{tab:dataset-stats}
\footnotesize
\begin{tabular}{@{}ll@{}}
\toprule
\textbf{Statistic} & \textbf{Value} \\ \midrule
Workflows & 500 \\
Fallback depth & 0--3 (125 each) \\
Path length & 11--38 steps \\
Goal length & $\sim$929 words \\
\bottomrule
\end{tabular}

\end{wraptable}

We construct an evaluation set of 500 workflows and execution paths, stratified evenly across fallback depths (Table~\ref{tab:dataset-stats}). Workflow complexity varies along two dimensions: \textbf{path length}, the total number of steps attempted from start to successful completion, and \textbf{fallback depth}, the number of tasks whose primary method fails completely and forces a switch.

We build the workflow tasks on the MultiWOZ domains \citep{budzianowski2020multiwozlargescalemultidomain}---hotel, train, taxi, restaurant, and attraction---each of which supplies realistic booking scenarios with natural constraints such as time slots and availability. We extend beyond single-domain dialogues to workflows whose tasks must be completed in sequence under cross-task dependencies. Each goal is a fixed template of nested conditional instructions, so the simulator must work through the branch structure to successfully achieve the goal.

\begin{figure*}[t]
  \centering
  \includegraphics[width=\textwidth]{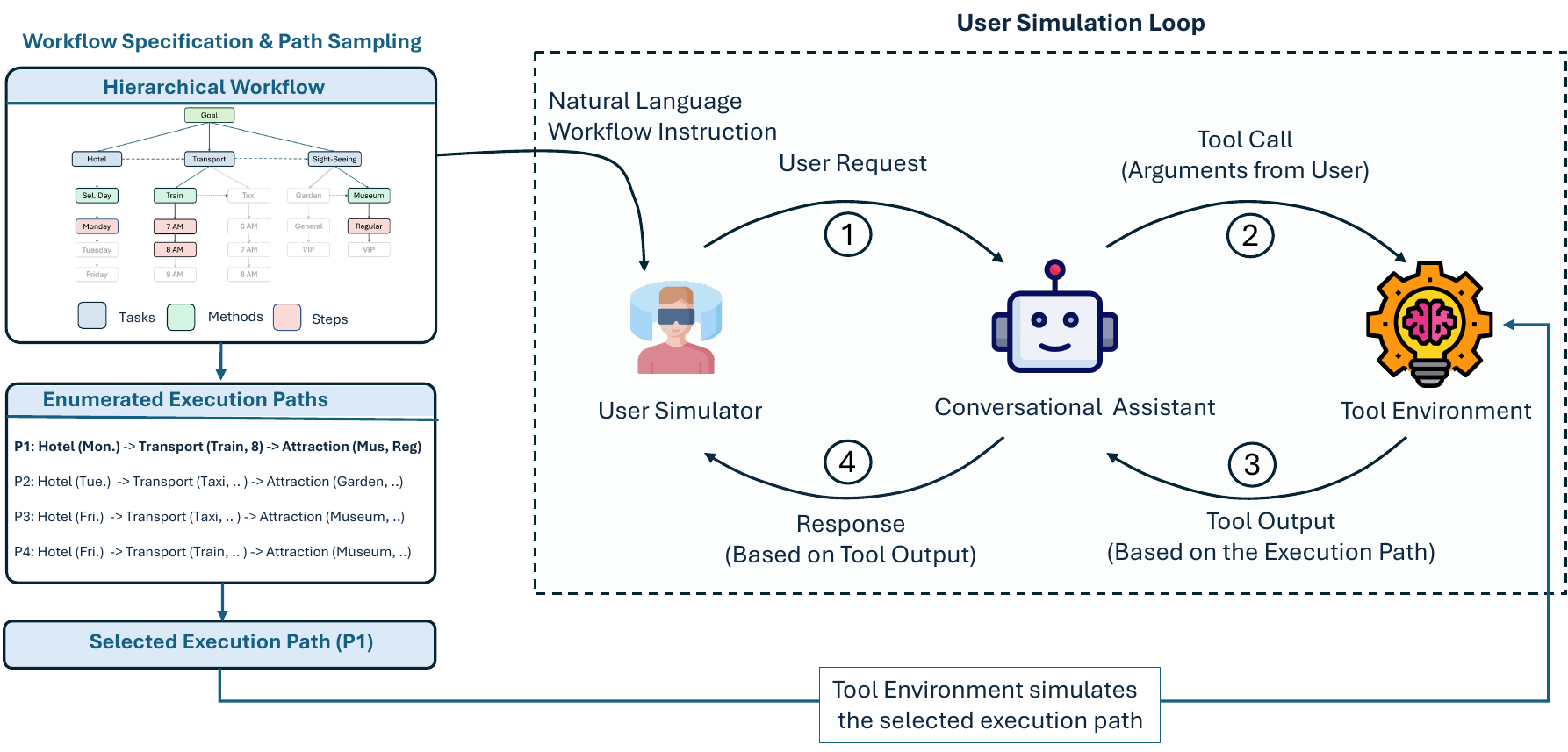}
  \caption{Evaluation framework. \textbf{Left:} workflows are specified over tasks, methods, and steps, and one execution path is sampled for evaluation. \textbf{Right:} the simulator receives natural-language workflow instructions and talks to an assistant, which makes tool calls to a controlled environment. The environment returns scripted outputs from an execution path \textbf{unknown} to the simulator, isolating simulator behavior.}
  \label{fig:user-simulation}
\end{figure*}

\subsection{Data Generation Pipeline}

Generation runs in three stages. First, we define HTN workflows by sampling tasks and their hierarchy, specifying primary and fallback methods with ordered steps, and sample an execution path from each. Second, for each path we generate the expected behavior at every step: the exact tool call, the argument values, and the correct decision (retry, fallback, or advance). Third, we convert the workflow structure into natural-language instructions covering the task sequence, available methods, ordered steps, and cross-task dependencies. This conversion is rule-based and deterministic, so the same workflow always produces the same goal text. The simulator sees only this text, never the underlying HTN or the sampled path.

\subsection{Controlled Conversation Loop}

The loop runs between the simulator, a conversational assistant, and a scripted environment (Figure~\ref{fig:user-simulation}): the simulator generates natural-language requests, the assistant parses them and executes tool calls, and the environment returns scripted success or failure responses based on the chosen evaluation path. Because the environment fully determines which steps succeed, deviations from ground truth originate almost entirely with the simulator, through incorrect arguments, wrong task selection, bad ordering, or premature fallback switches. The assistant can also err, but a reference-free LLM judge found assistant argument error below 0.4\% (\S \ref{sec:error-analysis}).

\section{Evaluation Metrics}
\label{sec:metrics}

A simulator can fail in several ways: calling the wrong tool, supplying incorrect arguments, or departing from the prescribed order of steps. \textbf{Tool Accuracy} is the fraction of steps where the correct tool was selected, and \textbf{Argument Accuracy} the fraction where every argument matches ground truth; together they separate navigation errors from argument errors. \textbf{Survival Rate} is the fraction of the workflow completed before the first error, which distinguishes early cascading failures from near-successes. Our primary metric, \textbf{Goal Success Rate (GSR)} \citep{davidson2023usersimulationlargelanguage}, asks whether the simulator executed the whole workflow exactly as prescribed: every expected call, in order, with correct arguments. Formal definitions are in Appendix~\ref{sec:eval-metrics}.

\section{Experiments}

We evaluate user simulator performance on our 500-workflow stratified evaluation set (125 workflows at each fallback depth 0--3) along two axes: the simulator's prompting strategy and the assistant's persona. We focus on open-weight models, which are cost-effective at scale and can be self-hosted. Frontier models would likely score higher but are expensive to run. Our aim is to understand the failure modes of accessible models and to find mitigations that make simulation with them practical. We set temperature to zero for both the user simulator and the assistant, and retry only on transient infrastructure failures (Appendix~\ref{sec:retry}).

\subsection{Experimental Conditions}

\subsubsection{Simulator Models}
We run four open-weight models \citep{grattafiori2024llama3herdmodels,deepseekai2025deepseekv3technicalreport,yang2025qwen3technicalreport} from 32B to 671B parameters, grouped into Small (\qwenthirtytwo), Medium (\llamaseventy and Qwen3 235B), and Large (DeepSeek V3.1 671B) tiers. The assistant is held fixed for all experiments.\footnote{we use Nova Pro \citep{agi2025amazonnovafamilymodels}}

\subsubsection{Assistant Personas}
\label{sec:personas}
We evaluate each simulator against five assistant personas that differ only in response style. The personas are: \emph{Terse} responds with success or failure; \emph{Tool-only} names the task that succeeded or failed; \emph{Echo} restates the parameters of the current attempt; and \emph{History} lists every value tried so far for that tool. The fifth persona, \emph{Suggestive}, replies like \emph{Echo} but after a failure volunteers one or two alternatives outside the goal, testing whether the simulator holds its prescribed ordering under unsolicited advice. We sweep these personas to quantify how much of a simulator's measured fidelity is attributable to the assistant's response style rather than to the simulator itself.

\subsubsection{Prompting Strategies}
\label{sec:ablations}

Completing a workflow requires the simulator to remember which options it has already tried and which task it is on. We therefore evaluate five strategies that differ in what the simulator is asked to track across turns, from nothing at all to structured state with reasoning over it.

\begin{enumerate}[leftmargin=1.4em,itemsep=1pt,topsep=2pt,parsep=0pt]
\item \textbf{Baseline} gives the workflow goal as natural-language instructions and asks the simulator to generate user utterances turn by turn, with no reasoning or state-tracking support.
\item \textbf{Goal Reminder} repeats the full goal at the end of each turn's prompt, testing whether restating it without any structure is enough.
\item \textbf{Summary} asks the simulator to maintain three fields across turns: a cumulative free-text summary of every task attempted so far and its outcome, the current task, and the next action.
\item \textbf{State Tracking Only} replaces the summary with a structured JSON record (current task, fixed parameters, variable options, tried and remaining options), testing whether enumerating the option space suffices without reasoning over it.
\item \textbf{ReUST} (Reasoning + User State Tracking) reasons over that state. At each turn the simulator receives the goal, the conversation so far, and the previous turn's state, then emits a JSON block recording the current task, which steps it has attempted and how they turned out, and the next action to take. It generates its utterance from that block.
\end{enumerate}

The instructions are task-agnostic; the Baseline and ReUST prompts, and an example of the state ReUST produces, are shown in Appendix~\ref{sec:prompts}.

\section{Results}

\subsection{Assistant Persona as a Confounder}
\label{sec:persona-sweep}

\begin{figure}[t]
\centering
\includegraphics[width=\linewidth]{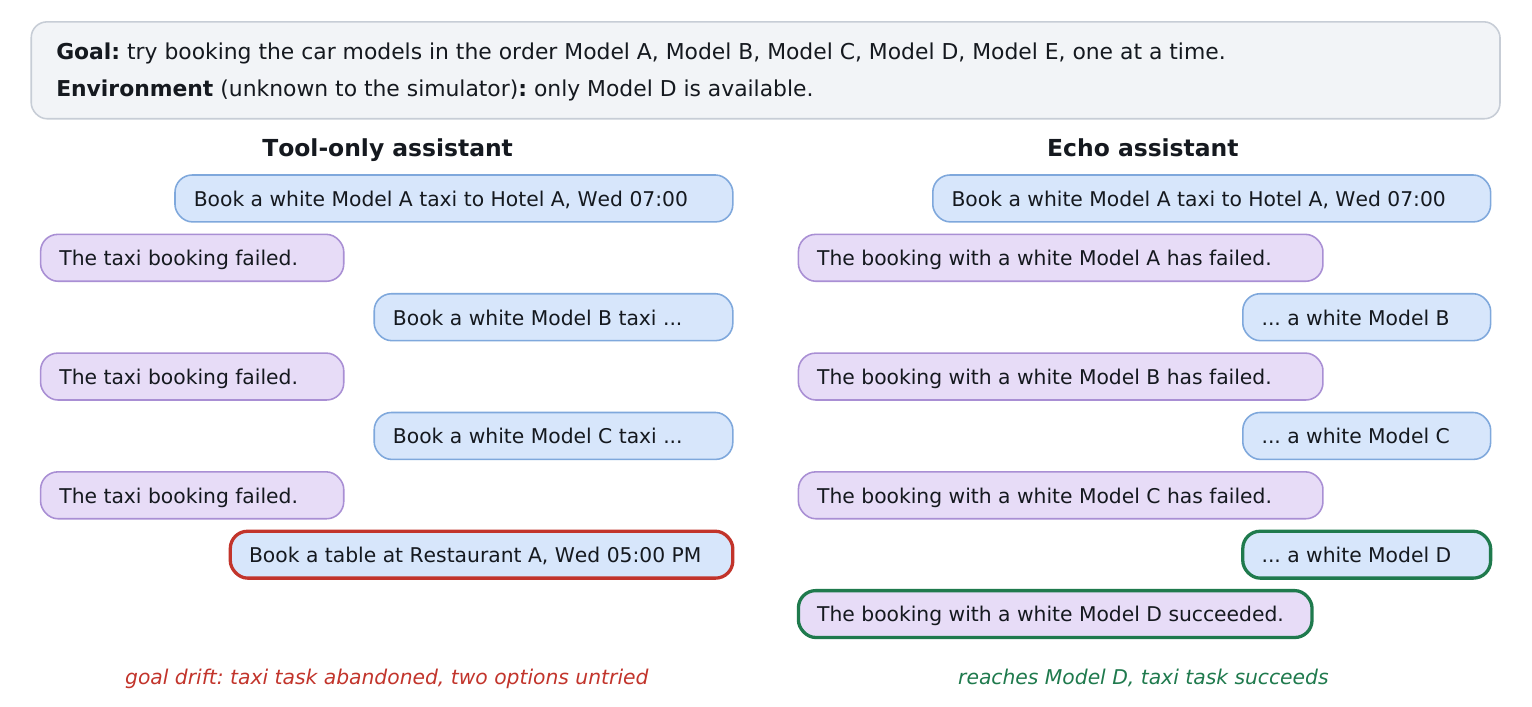}
\caption{The same simulator on the same workflow under two assistants, with venue and car model names replaced by placeholders. Under \emph{Tool-only} the simulator abandons the taxi task after three of the five models and moves on to the restaurant, which fails the workflow. Under \emph{Echo}, where the assistant repeats the model it just tried, the simulator reaches the fourth model and the task succeeds. 82.8\% of the workflows that fail under \emph{Tool-only} pass under \emph{Echo}.}
\label{fig:persona-example}
\end{figure}

We ran the same simulator against all five assistant personas under identical experimental settings. When the assistant uses the \emph{Tool-only} persona, which replies with only whether the task succeeded or failed, the simulator completes only 62.8\% of workflows correctly. However, when the assistant uses \emph{Echo}, which restates the details of the attempt it just made, the same simulator completes 92.8\% of workflows (Table~\ref{tab:ablation-results}, right). The assistant's reply acts as an external memory, so the simulator no longer has to track what it has already tried, and it achieves higher GSR as a result (Figure~\ref{fig:persona-example}). User simulator fidelity should therefore be established by testing against several assistant response styles. For the remaining experiments we use the \emph{Terse} assistant unless we say otherwise.

We then ask whether ReUST removes that dependence. Under ReUST, the standard deviation of GSR across personas falls from 12.4 pt to 2.0 pt (Table~\ref{tab:ablation-results}, right). Once the simulator tracks its own history, the choice of assistant persona has a small effect on what we measure.

\begin{table}[t]
\centering
\caption{GSR (\%) on Qwen3 235B. \textbf{Left:} prompting-strategy ablation under the default \emph{Terse} assistant. Goal Reminder tests goal salience alone; Summary tests lightweight NL state; State Tracking Only tests structured state without reasoning. \textbf{Right:} Baseline vs.\ ReUST under the five assistant personas of \S\ref{sec:personas}, in the order introduced there, with $\Delta$ the per-persona improvement; the Terse row is the same run as Baseline and ReUST on the left. The final row is the mean and sample standard deviation over the five personas, measuring the spread the persona axis induces rather than sampling error.}
\label{tab:ablation-results}
\begin{minipage}[t]{0.44\textwidth}
\centering
\small
\begin{tabular}{@{}lc@{}}
\toprule
\textbf{Strategy} & \textbf{GSR} \\ \midrule
Baseline            & 72.6 \\
Goal Reminder       & 64.2 \\
Summary             & 80.0 \\
State Tracking Only & 35.0 \\
ReUST               & 93.2 \\ \bottomrule
\end{tabular}
\end{minipage}\hfill
\begin{minipage}[t]{0.52\textwidth}
\centering
\small
\begin{tabular}{@{}lccc@{}}
\toprule
\textbf{Persona} & \textbf{Base.} & \textbf{ReUST} & \textbf{$\Delta$} \\ \midrule
Tool-only  & 62.8 & 96.0 & $+33.2$ \\
Terse      & 72.6 & 93.2 & $+20.6$ \\
Echo       & 92.8 & 94.2 & $+1.4$  \\
History    & 82.0 & 90.6 & $+8.6$  \\
Suggestive & 90.0 & 93.4 & $+3.4$  \\ \midrule
Mean $\pm$ SD & $80.0 \pm 12.4$ & $93.5 \pm 2.0$ & $+13.5$ \\ \bottomrule
\end{tabular}
\end{minipage}

\end{table}

\subsection{State Tracking Improves Fidelity}
\label{sec:ablation-study}

ReUST improves GSR from 81.0\% to 93.6\% on the Medium models and from 84.6\% to 97.4\% on the Large model (Table~\ref{tab:combined-results}). The increase comes primarily from argument accuracy on the Medium models (91.1\% to 97.6\%) in the taxi, restaurant, and attraction bookings, where the arguments have to be derived from the outcome of an earlier task (Appendix~\ref{sec:domain-breakdown}). We also compare ReUST with three simpler prompting strategies defined in \S\ref{sec:ablations} namely, repeating the goal every turn, summarizing the conversation in free text, and tracking state without reasoning. ReUST outperforms all three, reaching 93.2\% GSR (Table~\ref{tab:ablation-results}, left).

Each task in a workflow comes with a list of options to try in order, such as five car models for a taxi, and the task is finished as soon as one of them succeeds. State Tracking Only stores which options the simulator has tried and which are left, but not whether the last attempt succeeded. In half the runs (252 of 500), the simulator books the taxi and then carries on down the list, asking for the remaining car models. These extra calls violate strict goal following, which is why the strategy scores 35.0\%, below the 72.6\% of Baseline.

\begin{table*}[t]
\centering
\caption{Baseline vs.\ ReUST across all metrics (\%) on the 500-workflow stratified subsample. Tiers: Small = Qwen3 32B; Medium = Qwen3 235B and Llama 3.3 70B; Large = DeepSeek V3.1 671B. ReUST improves GSR by more than 12 points on Medium and Large, and lowers it on Small. 95\% confidence intervals are in Appendix~\ref{sec:confidence-intervals}.}
\label{tab:combined-results}
\small
\setlength{\tabcolsep}{5pt}
\begin{tabular}{@{}lcccccccc@{}}
\toprule
 & \multicolumn{2}{c}{\textbf{GSR}} & \multicolumn{2}{c}{\textbf{Tool Acc.}} & \multicolumn{2}{c}{\textbf{Arg. Acc.}} & \multicolumn{2}{c}{\textbf{Survival Rate}} \\
\textbf{Group} & \textbf{Baseline} & \textbf{ReUST} & \textbf{Baseline} & \textbf{ReUST} & \textbf{Baseline} & \textbf{ReUST} & \textbf{Baseline} & \textbf{ReUST} \\ \midrule
Small (32B)     & 25.2 & 16.8 & 85.0 & 79.4 & 69.9 & 74.1 & 60.7 & 58.8 \\
Medium (70--235B) & 81.0 & 93.6 & 95.7 & 97.8 & 91.1 & 97.6 & 89.5 & 97.4 \\
Large (671B)    & 84.6 & 97.4 & 99.8 & 99.4 & 97.0 & 99.2 & 93.0 & 97.7 \\ \bottomrule
\end{tabular}

\end{table*}

\subsection{Fidelity Degrades with Workflow Length}
\label{sec:complexity}

\begin{figure}[t]
    \centering
    \includegraphics[width=0.5\linewidth]{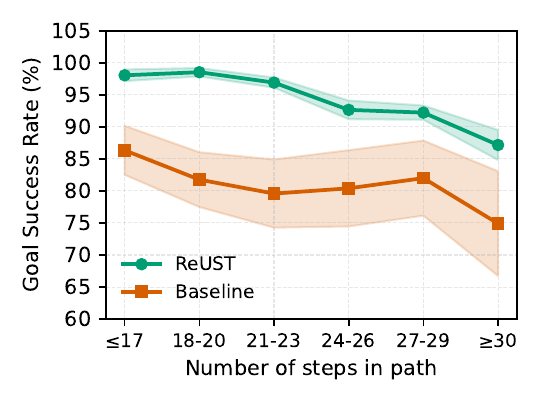}
    \caption{GSR vs path length on Qwen3 235B, averaged over the five assistant personas. Bands: $\pm$1 standard error across personas; each bin holds $\geq 40$ workflows.}
    \label{fig:gsr-vs-steps-reust}
\end{figure}

We use path length, the number of steps a workflow requires, as our measure of complexity. Longer paths ask the simulator for more values and give it more chances to lose track of what it has already done. GSR degrades with path length under both strategies, by a similar amount for each (Figure~\ref{fig:gsr-vs-steps-reust}). ReUST offers two advantages: it scores higher than Baseline at every path length, and its spread across assistant personas is much narrower, which makes the measurement less noisy.

\section{Error Analysis}
\label{sec:error-analysis}

To understand why simulators fail, we use an LLM judge to label every deviation from the workflow goal in the failed conversations. We analyze the Qwen3 235B runs under the default \emph{Terse} assistant, 137 and 34 failed workflows out of 500 for Baseline and ReUST respectively, using Claude Opus 4.5 \citep{anthropic2025claudeopus45} at temperature zero as the judge. The judge sees the goal and the full transcript, and assigns each deviation one of six failure modes, defined with examples in Appendix~\ref{sec:failure-modes}. Table~\ref{tab:failure-comparison} counts occurrences rather than conversations, since one conversation can show several modes.

\begin{table*}[t]
\centering
\caption{Failure mode occurrences over all failed conversations per condition (Qwen3 235B, \emph{Terse} assistant); a conversation may exhibit several modes. Significance: exact McNemar test on whether a workflow exhibits the mode at all, paired over the same 500 workflows, *$p < 0.05$, **$p < 0.001$.}
\label{tab:failure-comparison}
\small
\begin{tabular}{@{}lccc@{}}
\toprule
\textbf{Failure Mode} & \textbf{Baseline Count} & \textbf{ReUST Count} & \textbf{Relative Change} \\ \midrule
Goal Drift** & 152 & 22 & $-$85.5\% \\
Inter-Task Dependency Failure* & 16 & 3 & $-$81.2\% \\
Early Termination** & 136 & 34 & $-$75.0\% \\
Other (misinterpretation, incomplete arguments) & 0 & 2 & --- \\ \bottomrule
\end{tabular}

\end{table*}

Under Baseline, half of all failures are goal drift: the simulator skips a step, switches to a fallback too early, or attempts tasks out of order. ReUST reduces goal drift by 85.5\% in total, and also per failed conversation ($p < 0.001$ for both). Early termination is flagged in almost every failed conversation under both strategies, and its reduction follows the overall reduction in failures rather than being an independent effect. Using the same judge model and sub-sampled dataset, we find the assistant's tool selection accuracy and argument accuracy are 99.98\% and 99.67\% respectively, so the assistant is not a meaningful source of the GSR differences we report.

\section{Conclusion}

We introduced a 500-workflow HTN-based evaluation set for LLM-based user simulators, combining sequential execution, fallback logic, and cross-task dependencies. Our main finding is that user simulator fidelity depends on the assistant, so the assessment has to be done across several assistant response styles. Further, our proposed ReUST prompting strategy improves simulator fidelity across assistant personas for sufficiently capable models, making it a practical method. We also observe that, fidelity declines as workflows grow longer, so simulating long workflows faithfully is still a challenging problem.

\section*{Limitations}

The evaluation set covers five booking domains (hotel, train, taxi, restaurant, and attraction) within a single travel setting, so we cannot say how these results transfer to other task families. Our error analysis relies on LLM-as-a-Judge throughout, and we did not validate its labels against human annotation, so a judge that systematically mislabels a mode would bias the taxonomy in a direction we cannot detect. Our workflows are strictly forward-only---a completed task is never revisited and methods run in fixed order---whereas real users backtrack, so GSR measures fidelity to a prescribed policy rather than human-likeness.

\clearpage
\bibliography{custom}
\bibliographystyle{plainnat}

\clearpage
\appendix

\section{Evaluation Metrics}
\label{sec:eval-metrics}
\subsection{Tool Accuracy}

Tool Accuracy measures whether the simulator triggers the correct tool at each step, independent of argument values. This isolates navigation errors (e.g., calling \texttt{book\_taxi} when \texttt{book\_train} was expected) from argument errors (e.g., correct tool but wrong time). A simulator might select the right tool but provide wrong arguments, or vice versa---Tool Accuracy captures only the former.

Let $t_i^*$ be the expected tool at step $i$, $\hat{t}_i$ the simulator's tool, $N$ the number of expected steps, and $M$ the number of calls the conversation actually produced:
\[
\text{Tool Accuracy} = \frac{1}{\max(M, N)}\sum_{i=1}^{\min(M, N)} \mathbf{1}\!\left[\hat{t}_i = t_i^*\right]
\]
The $\max(M, N)$ denominator penalizes calls beyond the expected sequence. With $N$ alone, a simulator that finishes the workflow and then keeps requesting alternatives would score the same as one that stops on time.

\subsection{Argument Accuracy}

Argument Accuracy measures whether the simulator provides correct argument values when making requests to the assistant. Each tool call requires specific arguments (e.g., \texttt{book\_hotel(day="Monday", room\_type="suite", guests=5)}), and the ground-truth path specifies the expected values at each step. This metric captures errors such as providing wrong dates, incorrect counts, or values that violate cross-task dependencies (e.g., booking a taxi for Tuesday when the hotel was reserved for Monday).

A step is correct only if all its arguments match ground truth. Let $\mathbf{a}_i^*$ be the ground-truth arguments for step $i$ and $\hat{\mathbf{a}}_i$ be the simulator's arguments. Argument Accuracy averages across all $N$ steps:
\[
\text{Argument Accuracy} = \frac{1}{N}\sum_{i=1}^{N} \mathbf{1}\!\left[\hat{\mathbf{a}}_i = \mathbf{a}_i^*\right]
\]

We normalize the arguments before matching to match minor variations between ground truth and predicted args.

\subsection{Survival Rate}

Survival Rate measures how far into the workflow the simulator progresses before making its first error. Unlike Argument Accuracy which averages over all steps, Survival Rate captures whether errors occur early (causing cascading failures) or late (indicating near-success). For example, a simulator that correctly executes 8 of 10 steps before failing has Survival Rate 0.8, while one that fails on step 2 has Survival Rate 0.2.

Let $N$ be the total steps and $k$ be the number of correct steps before the first error. Then:
\[
\text{Survival Rate} = \frac{k}{N}
\]
A perfect trajectory yields 1.0; early failures yield proportionally lower scores.

\subsection{Goal Success Rate}

Goal Success Rate is our primary metric, measuring whether the simulator completed the entire workflow correctly. A goal is completed only if the conversation produces exactly as many calls as the ground-truth path expects and every one of them matches on both tool and arguments. Unlike Argument Accuracy (which can be high even with errors) or Survival Rate (which rewards partial progress), GSR is binary: a workflow either succeeds completely or fails. This metric reflects real-world requirements where a partially completed booking workflow (e.g., hotel reserved but taxi not booked) represents a failure.

The length condition rules out trajectories that satisfy the whole expected sequence and then continue. Survival Rate scores only the expected prefix, so it assigns those trajectories 1.0; GSR does not, and the two metrics disagree on exactly this case.

For workflow $i$ with $N_i$ expected steps and $M_i$ observed calls:
\[
\text{success}_i =
\begin{cases}
1 & \text{if } M_i = N_i \text{ and } \hat{t}_j = t_j^*,\, \hat{\mathbf{a}}_j = \mathbf{a}_j^* \;\forall j \leq N_i \\
0 & \text{otherwise}
\end{cases}
\]
Across $N$ workflows:
\[
\text{GSR} = \frac{1}{N}\sum_{i=1}^{N} \text{success}_i
\]

\section{Definition of Errors}
\label{sec:failure-modes}
\paragraph{Goal Drift} The user simulator deviates from the specified workflow sequence. This includes: (1) skipping intermediate steps within a task and jumping to future steps, (2) switching to a fallback method without exhausting the current method, (3) attempting tasks out of order, violating dependencies. \textit{Example (Intra-Task):} After a booking fails for Wednesday, the user should try Thursday next, but instead skips to Friday. \textit{Example (Inter-Task):} User tries to book a taxi to an attraction before booking the attraction itself.

\paragraph{Inter-Task Dependency Failure} The user simulator fails to maintain or correctly use information from previously completed tasks. Details from past tasks (e.g., booking reference, location name, time) are needed for future tasks but are not used or are used incorrectly. \textit{Example:} User books a restaurant but then fails to use the restaurant's location when booking a taxi to that restaurant.

\paragraph{Misinterpretation of Assistant Response} The user simulator fails to understand the assistant's response and makes a wrong decision. User retries a task that was already successful, or moves on when they should retry. \textit{Example:} Assistant confirms ``restaurant booking successful for Thursday at 07:00 PM'' but user tries to book again instead of moving to next task.

\paragraph{Early Termination} The user simulator ends the conversation before completing all tasks specified in the workflow goal. The conversation up to that point may be correct, but it is incomplete.

\paragraph{Incomplete Arguments} The user simulator omits preference arguments when making requests. Main arguments are provided but optional/preference arguments are missing. \textit{Example:} For train booking, user provides departure, destination, time but omits the AC-tier class preference.

\paragraph{Assistant Error} The assistant makes tool calls with incorrect arguments or to wrong tools, hallucinates tool results, or provides information not returned by the tool.

\section{Argument Accuracy by Domain}
\label{sec:domain-breakdown}

Table~\ref{tab:arg-accuracy-gain} breaks argument accuracy down by booking domain, pooled over the four simulators under the default \emph{Terse} assistant.

\begin{table}[H]
\centering
\caption{Argument accuracy by domain (\%) on the 500-workflow stratified evaluation set, pooled over the four simulators under the default \emph{Terse} assistant. A call counts as correct when the tool matches and every argument value matches, so each row is a per-call rate over that domain's ground-truth calls, with calls the simulator never reached counted as incorrect; the final row pools all calls and therefore differs slightly from the workflow-level means in Table~\ref{tab:combined-results}. ReUST yields the largest gains in Restaurant, Taxi, and Attraction, where arguments depend on prior task outcomes. Hotel and Train are at ceiling under both conditions.}
\label{tab:arg-accuracy-gain}
\small
\begin{tabular}{@{}lccc@{}}
\toprule
\textbf{Domain} & \textbf{Baseline} & \textbf{ReUST} & \textbf{$\Delta$} \\ \midrule
Hotel      & 100.0 & 99.9 & $-$0.1 \\
Taxi       & 82.6 & 89.2 & +6.6 \\
Attraction & 80.0 & 85.3 & +5.3 \\
Restaurant & 82.9 & 92.4 & +9.5 \\
Train      & 98.6 & 98.6 & $-$0.0 \\ \midrule
All domains & 86.4 & 91.6 & +5.2 \\ \bottomrule
\end{tabular}

\end{table}

\section{Retry Policy}
\label{sec:retry}

The retry mechanism operates at two levels: (1) API-level retries (up to 10 attempts with exponential backoff, capped at 30 seconds) handle transient infrastructure failures such as rate limiting and API timeouts; (2) conversation-level retries (up to 5 attempts) restart the entire conversation if all API retries are exhausted. Under deterministic decoding, successful API calls produce identical outputs, so retries do not affect semantic behavior. The retry configuration is identical across all prompting strategies.

\section{Prompts}
\label{sec:prompts}

\begin{tcolorbox}[
    title={Assistant Prompt},
    fonttitle=\bfseries,
    center title
]
\small\ttfamily

You are a Trip Booking Assistant.\\[4pt]

<instructions>\\
Follow this exact process for every user request:\\[2pt]

1. \textbf{Parameter Check}: Examine if ALL required parameters for the booking tool are provided.\\[4pt]

2. \textbf{Decision}:\\
\hspace*{1em}(i) If ANY parameter is missing $\rightarrow$ Ask for providing all the parameters (do not proceed). Always ask for clarification when parameters are missing.\\
\hspace*{1em}(ii) If ALL parameters are complete $\rightarrow$ Make the tool call and convey the results to the user.\\[4pt]

3. \textbf{Response}: If you get a response from the tool call, you need to clearly summarize the output and respond to the user.\\[4pt]

</instructions>\\[6pt]

<constraints>\\
(i) NEVER invent, assume, or hallucinate missing parameter values.\\
(ii) NEVER make tool calls with incomplete information.\\
(iii) ALWAYS ask for clarification when parameters are missing.\\
</constraints>\\[6pt]

<formatting>\\
(i) Be direct and concise.\\
(ii) Single tool call at a time.\\
(iii) Clearly communicate booking results (success/failure) you receive from the tool call.\\
(iv) When asking for missing parameters, list them clearly.\\
</formatting>

\end{tcolorbox}

\vspace{12pt}

\begin{tcolorbox}[
    title={User Simulator Zero Shot Prompt},
    fonttitle=\bfseries,
    center title,
    breakable
]
\small\ttfamily

You are a user interacting with an agent for booking a trip. \\[4pt]

<rules>\\[2pt]

<conversation\_flow>\\
* Wait for agent response before trying next alternative if current option unavailable\\
* Ask for ONE option at a time -- do NOT bundle alternatives\\
* Generate one line at a time and be specific with every detail\\
* Do not reveal all instructions at once -- only provide information necessary for the current step\\
</conversation\_flow>\\[4pt]

<content\_rules>\\
* NEVER give conditional statements or reference previous bookings -- ask directly for what you need\\
* Never forget to provide the required values for the current task\\
* Be direct and specific -- no background information or context from previous steps\\
* Make conversation natural and never provide reasoning or explanations\\
</content\_rules>\\[4pt]

<communication\_style>\\
* NEVER say things like 'The restaurant booking is at...' or 'Since the hotel is...' -- just make direct requests\\
* Simulate natural user messages with explicit and exact details\\
</communication\_style>\\[4pt]

<termination>\\
* IMPORTANT: When ALL tasks are completed successfully, generate \texttt{\#\#\#STOP\#\#\#} as a standalone message without anything else to end the conversation\\
</termination>\\[4pt]

</rules>\\[6pt]

<goal>\\
{[}WORKFLOW INSTRUCTIONS HERE{]}\\
</goal>\\[6pt]

Conversation so far:\\
Next user utterance:\\[2pt]

\end{tcolorbox}

\clearpage

\begin{tcolorbox}[
    title={ReUST Prompt},
    fonttitle=\bfseries,
    center title,
    breakable
]
\small\ttfamily

You are a user interacting with an agent for booking a trip.\\[6pt]

<response\_format>\\
First, generate a Thought as a structured JSON-like format showing:\\
- Current task and its options (fixed parameters vs variable options to try)\\
- What has been tried so far\\
- What options are left to try\\
- Next action to take\\
(This thought will not be sent to the agent)\\[4pt]

Then, generate a one line User Response to simulate the user's message (this message will be sent to the agent).\\[4pt]

Format:\\
Thought: \{\\
\hspace*{1em}"current\_task": "<task\_name>",\\
\hspace*{1em}"previous\_turn\_summary": "<what\_was\_tried\_and\_outcome>",\\
\hspace*{1em}"fixed\_params": \{"<param>": "<value>", ...\},\\
\hspace*{1em}"variable\_options": \{"<param>": ["option1", "option2", ...]\},\\
\hspace*{1em}"tried": ["option1", ...],\\
\hspace*{1em}"remaining": ["option2", ...],\\
\hspace*{1em}"next\_action": "<what\_to\_do\_next>"\\
\}\\
User Response: [The actual message to send to the agent]\\[6pt]
</response\_format>\\[8pt]

<rules>\\[4pt]

<conversation\_flow>\\
* Wait for agent response before trying next alternative if current option unavailable\\
* Ask for ONE option at a time -- do NOT bundle alternatives\\
* When instructions say "retry consecutively/successively", try options in sequence one by one\\
* Generate one line at a time and be specific with every detail\\
* Do not reveal all instructions at once -- only provide information necessary for the current step\\
</conversation\_flow>\\[6pt]

<content\_rules>\\
* NEVER give conditional statements or reference previous bookings -- ask directly for what you need\\
* Never forget to provide the required values for the current task\\
* Be direct and specific -- no background information or context from previous steps\\
* Make conversation natural and never provide reasoning or explanations in the User Response\\
</content\_rules>\\[6pt]

<communication\_style>\\
* NEVER say things like 'The restaurant booking is at...' or 'Since the hotel is...' -- just make direct requests\\
* Simulate natural user messages with explicit and exact details\\
</communication\_style>\\[6pt]

<termination>\\
* IMPORTANT: When ALL tasks are completed successfully, generate \texttt{\#\#\#STOP\#\#\#} as a standalone message without anything else to end the conversation\\
</termination>\\[4pt]

</rules>

\end{tcolorbox}

\clearpage

\subsection{Example ReUST State}
\label{sec:reust-state-example}

In the blocks below, we show two consecutive turns for the default \emph{Terse} assistant. At turn~2 the simulator has completed the hotel task and records the train options it has yet to try; the 06:00~AM request then fails, and at turn~3 that option has moved from \texttt{remaining} to \texttt{tried} and \texttt{next\_action} advances to the next option in the prescribed order.

\begin{tcolorbox}[
    title={ReUST State, Turn 2},
    fonttitle=\bfseries,
    center title,
    breakable
]
\footnotesize
\begin{verbatim}
{
  "current_task": "TRAVEL_BOOKING",
  "previous_turn_summary": "Successfully booked deluxe room at Holiday
    Inn for 5 people for 2 nights starting on Tuesday.",
  "fixed_params": {
    "departure": "London Victoria",
    "destination": "Cambridge",
    "number_of_people": 5,
    "travel_day": "Tuesday"
  },
  "variable_options": {
    "train_time": ["06:00 AM", "07:00 AM", "08:00 AM", "09:00 AM"]
  },
  "tried": [],
  "remaining": ["06:00 AM", "07:00 AM", "08:00 AM", "09:00 AM"],
  "next_action": "Ask for train ticket at 06:00 AM"
}
\end{verbatim}
\textbf{Utterance:} I would like to book Ac tier-class train tickets from London Victoria to Cambridge for 5 people on Tuesday at 06:00 AM.
\end{tcolorbox}

\begin{tcolorbox}[
    title={ReUST State, Turn 3},
    fonttitle=\bfseries,
    center title,
    breakable
]
\footnotesize
\begin{verbatim}
{
  "current_task": "TRAVEL_BOOKING",
  "previous_turn_summary": "Train booking failed for 06:00 AM on Tuesday
    from London Victoria to Cambridge for 5 people.",
  "fixed_params": {
    "departure": "London Victoria",
    "destination": "Cambridge",
    "number_of_people": 5,
    "travel_day": "Tuesday"
  },
  "variable_options": {
    "train_time": ["06:00 AM", "07:00 AM", "08:00 AM", "09:00 AM"]
  },
  "tried": ["06:00 AM"],
  "remaining": ["07:00 AM", "08:00 AM", "09:00 AM"],
  "next_action": "Ask for train ticket at 07:00 AM"
}
\end{verbatim}
\textbf{Utterance:} I would like to book Ac tier-class train tickets from London Victoria to Cambridge for 5 people on Tuesday at 07:00 AM.
\end{tcolorbox}

\clearpage

\section{Confidence Intervals}
\label{sec:confidence-intervals}

Table~\ref{tab:ci} reports 95\% confidence intervals for all metrics. The Medium tier (two models) uses a hierarchical bootstrap: resample models within the tier with replacement, then resample workflows within each model. The Small and Large tiers each contain a single model, so their CIs are computed via workflow-level bootstrap only. All bootstraps use 1{,}000 iterations.

\begin{table*}[h]
\centering
\caption{95\% confidence intervals for all metrics (\%) on the 500-workflow stratified evaluation set. Values reported as mean $\pm$ margin. The Medium tier uses a hierarchical bootstrap (resample models within tier, then workflows within model); Small and Large tiers contain a single model and use a workflow-level bootstrap. 1{,}000 iterations.}
\label{tab:ci}
\scriptsize
\setlength{\tabcolsep}{4pt}
\begin{tabular}{@{}lcccccccc@{}}
\toprule
 & \multicolumn{2}{c}{\textbf{GSR}} & \multicolumn{2}{c}{\textbf{Tool Acc.}} & \multicolumn{2}{c}{\textbf{Arg. Acc.}} & \multicolumn{2}{c}{\textbf{Survival Rate}} \\
\textbf{Group} & \textbf{Baseline} & \textbf{ReUST} & \textbf{Baseline} & \textbf{ReUST} & \textbf{Baseline} & \textbf{ReUST} & \textbf{Baseline} & \textbf{ReUST} \\ \midrule
Small (32B)     & $25.2 \pm 3.8$ & $16.8 \pm 3.3$ & $85.0 \pm 1.8$ & $79.4 \pm 1.7$ & $69.9 \pm 2.2$ & $74.1 \pm 2.3$ & $60.7 \pm 2.3$ & $58.8 \pm 2.8$ \\
Medium (70--235B) & $81.0 \pm 9.9$ & $93.6 \pm 1.7$ & $95.7 \pm 4.1$ & $97.8 \pm 1.0$ & $91.1 \pm 7.6$ & $97.6 \pm 1.0$ & $89.5 \pm 7.0$ & $97.4 \pm 1.2$ \\
Large (671B)    & $84.6 \pm 3.1$ & $97.4 \pm 1.5$ & $99.8 \pm 0.3$ & $99.4 \pm 0.6$ & $97.0 \pm 1.0$ & $99.2 \pm 0.7$ & $93.0 \pm 1.6$ & $97.7 \pm 1.2$ \\ \bottomrule
\end{tabular}

\end{table*}

\section{Example User Simulator Goal}
\label{sec:example-goal}

\begin{goalboxwide}
{\setlength{\parindent}{0pt}\small\ttfamily

Goal: Complete these booking tasks in sequence.\\[4pt]

<task\_sequence>\\[2pt]

<HOTEL\_BOOKING>\\
Book a suite room at Hotel A for 4 people for 3 nights. Your first preference is to book
the hotel on Monday; if a room is not available on Monday, retry for the next day (Tuesday),
continuing successively up to Wednesday. Provide your preference to the agent one by one
(e.g., first ask for Monday, then Tuesday and so on). Complete the remaining tasks using the
day on which the hotel was successfully reserved.\\[4pt]
</HOTEL\_BOOKING>\\[6pt]

Proceed to Travel Booking only if Hotel booking was successful.\\[4pt]

<TRAVEL\_BOOKING>\\
Book Ac tier-class train tickets from London St Pancras to Cambridge for 4 people on the same
day the hotel was successfully booked. Your first preference is to book the train leaving at
06:00 AM; if the train is not available at 06:00 AM, then retry for the next hour (07:00 AM),
continuing at hourly increments until 09:00 AM (inclusive). Provide your preference to the
agent one by one (e.g., first ask for 06:00 AM, then 07:00 AM and so on).\\[4pt]

After checking all hourly slots up to 09:00 AM, if no trains are available, book a taxi
instead. First ask for a sedan taxi from London St Pancras to Cambridge station at
06:00 AM on the hotel day; if the taxi is not available, ask for a sedan taxi at
07:00 AM on the hotel day, continuing until 08:00 AM.\\[4pt]
</TRAVEL\_BOOKING>\\[6pt]

Proceed to Taxi booking from station to hotel only if Travel Booking was successful.\\[4pt]

<TAXI\_FROM\_STATION\_TO\_HOTEL>\\
Book a taxi from Cambridge Station to Hotel A on the same day the hotel was successfully
booked. The car model preference order is:\\
\hspace*{1em}sedan type A\\
\hspace*{1em}sedan type B\\
\hspace*{1em}luxury sedan type A\\[4pt]

Use the following rule for choosing the taxi time:\\
If a train was successfully booked: book the taxi exactly 1 hour after the train's successful
booking time.\\
If no train was available and a taxi from London St Pancras to Cambridge Station was
successfully booked instead: book the taxi exactly 2 hours after the taxi's successful
booking time.\\[4pt]

At the chosen time, ask first for a white sedan type A. If not available, ask next for a white
sedan type B. If still not available, ask for a white luxury sedan type A. Ask one preference at a
time.\\[4pt]
</TAXI\_FROM\_STATION\_TO\_HOTEL>\\[6pt]

Proceed to Restaurant booking only if Taxi booking from station to hotel was successful.\\[4pt]

<RESTAURANT\_BOOKING>\\
Book a table at Restaurant A for 4 people. Your first preference is to book the table at
05:00 PM; if a table is not available at 05:00 PM, then retry for the next hour (06:00 PM),
continuing at hourly increments until 08:00 PM (inclusive). Provide your preference to the
agent one by one (e.g., first ask for 05:00 PM, then 06:00 PM and so on).\\[4pt]

After checking all time slots at Restaurant A, if no tables are available, book at Restaurant B
instead. First ask for Restaurant B at 05:00 PM; if not available, retry for the next hour
(06:00 PM), continuing at hourly increments until 07:00 PM. Remember the successful
restaurant and its booking time because the taxi between the hotel and restaurant depends
on them.\\[4pt]
</RESTAURANT\_BOOKING>\\[6pt]

Proceed to Taxi booking from hotel to restaurant only if Restaurant booking was successful.\\[4pt]

<TAXI\_FROM\_HOTEL\_TO\_RESTAURANT>\\
Book a taxi from Hotel A to the successfully booked restaurant on the same day the hotel
was successfully booked. The car model preference order is blue sedan type B, then blue
luxury sedan type A. Your first preference is to book a blue sedan type B taxi for exactly
30 minutes before the successful restaurant booking time; if a blue sedan type B taxi is not
available, retry with the next car model in the preference order at the same time. Provide
your preference to the agent one by one (e.g., first ask for blue sedan type B, then
blue luxury sedan type A and so on).\\[4pt]
</TAXI\_FROM\_HOTEL\_TO\_RESTAURANT>\\[6pt]

Proceed to Attraction booking only if Taxi booking from hotel to restaurant was successful.\\[4pt]

<ATTRACTION\_BOOKING>\\
Book Historic Site A tickets for 4 people on the day after the successful hotel booking. The
ticket access preference order is general access, standard access, luxury access, vip access.
Your first preference is to book general access tickets; if general access tickets are not
available, retry with the next access type on the same day, following the given preference
order. Provide your preference to the agent one by one (e.g., first ask for general access,
then standard access and so on).\\[4pt]

After checking all access types at Historic Site A, if no tickets are available, book at
Attraction B on the day after the successful hotel booking instead. The access type
preference order is: general access, standard access, luxury access.\\[4pt]
</ATTRACTION\_BOOKING>\\[6pt]

Proceed to Taxi booking from hotel to attraction only if Attraction booking was successful.\\[4pt]

<TAXI\_FROM\_HOTEL\_TO\_ATTRACTION>\\
Book a taxi from Hotel A to the attraction for which tickets were successfully booked,
scheduled at 11:00 AM on the same day as that attraction booking. The car model preference
order is black luxury sedan type A, black luxury sedan type B, black economy sedan type A, black economy sedan type B. Your
first preference is to book a black luxury sedan type A taxi; if a black luxury sedan type A taxi is not
available, retry with the next car model in the preference order at the same time. Provide
your preference to the agent one by one (e.g., first ask for black luxury sedan type A, then black
luxury sedan type B and so on).\\[4pt]
</TAXI\_FROM\_HOTEL\_TO\_ATTRACTION>\\[6pt]

Once all the tasks are completed, generate \#\#\#STOP\#\#\# as a standalone message without
anything else to end the conversation.\\[2pt]

</task\_sequence>

}
\end{goalboxwide}

\end{document}